\documentclass[11pt]{article}

\usepackage[letterpaper,margin=0.9in]{geometry}
\usepackage{times}            % Times-like text; remove for Computer Modern
\usepackage[T1]{fontenc}
\usepackage[utf8]{inputenc}
\usepackage{microtype}
\usepackage{graphicx}
\usepackage{booktabs}
\usepackage{multirow}
\usepackage{amsmath}
\usepackage{amssymb}          % \checkmark
\usepackage{xcolor}
\usepackage[numbers,sort&compress]{natbib}   % numbered [1], [1-4] citations
\usepackage{hyperref}
\hypersetup{colorlinks=true, citecolor=blue, linkcolor=blue, urlcolor=blue}
\usepackage{float}
\usepackage{enumitem}
\usepackage[skip=3pt plus1pt]{parskip}  % no first-line indent; small gap between paragraphs
\usepackage{tcolorbox}
\tcbuselibrary{breakable}

\graphicspath{{../figures/}{./figures/}}

\tcbset{promptstyle/.style={
  breakable,
  colback=white, colframe=black,
  colbacktitle=white, coltitle=black,
  fonttitle=\bfseries\footnotesize, fontupper=\footnotesize,
  boxrule=0pt,
  toprule=0.8pt, bottomrule=0.8pt,
  titlerule=0.4pt,
  arc=0pt,
  left=4pt, right=4pt, top=2pt, bottom=2pt,
  toptitle=1pt, bottomtitle=1pt,
  before skip=3pt, after skip=3pt
}}

\newcommand{\mds}{MemoryDocDataSet}

\title{\textbf{\mds: A Benchmark for Joint Conversational Memory and Long Document Reasoning}}

\author{
  Qiyang Xie$^{1}$ \quad Jialun Wu$^{2}$ \quad Xinjie He$^{3}$ \quad Su Liu$^{4}$ \\[3pt]
  Shuai Xiao$^{4}$ \quad Zhiyuan Lin$^{4}$ \quad Weikai Zhou$^{4}$ \\[6pt]
  $^{1}$Northeastern University \quad $^{2}$Johns Hopkins University \\[2pt]
  $^{3}$Columbia University \quad $^{4}$Independent Researcher
}
\date{\ifcase\month\or January\or February\or March\or April\or May\or June\or
  July\or August\or September\or October\or November\or December\fi\space\the\year}

\begin{document}
\maketitle

% ─────────────────────────────────────────────────────────────────────────────
\begin{abstract}
AI systems increasingly need to combine two demanding capabilities: navigating
multi-session conversation history and performing deep reading comprehension
within long documents. Yet no existing benchmark evaluates both simultaneously.
We introduce \textbf{\mds}, a synthetic benchmark of 50 micro-worlds and 1,000
QA pairs in which each instance comprises 3--5 personas, a temporal event graph
spanning months of activity, 3--5 real long documents (20,000--50,000 tokens
each sourced from the Caselaw Access Project), multi-session conversations
grounded on those documents, and 20 question-answer pairs across five reasoning
categories. The defining feature is the \textit{Hybrid} source tag: questions
requiring a system to first navigate conversation history to identify which
document is relevant, then extract the answer from within that document. Hybrid
questions account for 75.1\% of the dataset. Dataset quality is characterised through a prompt-sensitivity self-consistency
analysis using LLM-as-judge, yielding a median Cohen's $\kappa = 0.634$ across
all 50 micro-worlds. We
evaluate six baseline configurations spanning truncated context, long-context
LLMs, retrieval-augmented generation (RAG) \citep{lewis2020retrieval}, and
memory systems. The best baseline (RAG-Both) achieves 0.358 overall F1 and
0.342 on Hybrid. Document-only retrieval (RAG-Doc) collapses to 0.267 on
Hybrid despite achieving 0.453 on Doc-only questions, demonstrating a clear
joint-retrieval gap that motivates architectures unifying conversational memory
with long-document navigation. We release the dataset, generation pipeline, and
all baseline implementations.
\end{abstract}

% ─────────────────────────────────────────────────────────────────────────────
\section{Introduction}

Modern AI assistants operate in settings that demand two capabilities
simultaneously. First, they must maintain coherent memory across long
conversations---tracking who said what, when, and how prior context informs the
current turn. Second, they must perform deep reading comprehension within long
documents---contracts, reports, filings---that far exceed the size of any
conversational exchange.

These two capabilities have been studied largely in isolation. Memory benchmarks
such as LoCoMo \citep{maharana2024evaluating} and LongMemEval
\citep{wu2025longmemeval} present systems with multi-session dialogues but attach
no long documents; a system that memorizes conversation facts perfectly can score
highly. Long-document benchmarks such as L-Eval \citep{an2024leval} and
ZeroSCROLLS \citep{shaham2023zeroscrolls} present systems with book-length or
report-length texts but no conversation structure; a system with a large context
window can score highly without any notion of memory.

Neither family of benchmarks measures the combination. Yet real-world deployment
routinely requires it. Consider a legal assistant working with a client over
several months: it must recall from prior sessions which contract the client is
asking about, then navigate 40,000 tokens of that contract to answer a specific
question about a penalty clause. This task is trivial for neither a pure memory
system nor a pure long-context reader.

We introduce \textbf{\mds}, a benchmark dataset designed precisely to test this
joint capability. Each instance is a \textit{micro-world}---a self-contained
scenario comprising 3--5 \textbf{personas} with defined roles, expertise, and
relationships; a \textbf{temporal event graph} of 5--10 time-stamped events
spanning at least six months; 3--5 \textbf{long documents} of 20,000--50,000
tokens each sourced from real public domain legal corpora; five
\textbf{multi-session conversations} grounded on the event graph in which
speakers naturally reference the attached documents; and \textbf{20 QA pairs}
per micro-world spanning five reasoning categories.

The key novelty is the \textit{source dimension}---every QA pair is tagged with
whether its answer lies in the conversation history only (\textit{Chat-only}), in
a long document only (\textit{Doc-only}), or requires bridging both
(\textit{Hybrid}). Hybrid questions require a system to first use conversation
context to identify the relevant document, then read that document deeply to
extract the answer. We require at least 30\% of QA pairs per micro-world to be
Hybrid.

We evaluate six baseline configurations against our benchmark: a
truncated-context base LLM, a long-context LLM with full context, RAG over
conversations only, RAG over documents only, RAG over both, and a memory system
baseline. Results show that all six configurations underperform on Hybrid
questions, with RAG over both achieving the highest F1 but still far below
expected human performance (discussed in Section~\ref{sec:discussion-human}).
This gap motivates new systems that tightly integrate conversational memory with
long-document retrieval.

\paragraph{Contributions.}
\begin{enumerate}
  \item A benchmark dataset with 50 micro-worlds, 1,000 QA pairs, and a novel
        source-dimension annotation that no existing benchmark provides.
  \item A fully automated, config-driven pipeline for generating micro-worlds at
        scale using real long documents and LLM-generated synthetic structure.
  \item Baseline results across six retrieval and memory configurations,
        establishing the first systematic comparison on joint conversational
        memory and long-document reasoning.
\end{enumerate}

% ─────────────────────────────────────────────────────────────────────────────
\section{Related Work}

\subsection{Conversational Memory Benchmarks}

Several benchmarks evaluate long-term conversational memory. The Beyond Goldfish
Memory paper \citep{xu2022beyond}, which introduced the Multi-Session Chat (MSC)
dataset, pioneered multi-session dialogue evaluation but focuses on persona
consistency over short ($\sim$1K token) exchanges. LoCoMo
\citep{maharana2024evaluating} extends this to longer conversations (up to 300
turns, $\sim$9K tokens) and introduces temporal reasoning questions.
LongMemEval \citep{wu2025longmemeval} evaluates LLMs as conversation partners
across multiple sessions, with questions probing what was said in prior turns.
MemBench \citep{tan2025membench} provides a structured suite of memory operations
including storage, retrieval, and update.

A common limitation across all these benchmarks is the absence of long documents.
The context is entirely conversational; systems that memorize conversation facts
well can achieve high scores without any document reading capability.

\subsection{Long-Document and Multi-hop Benchmarks}

A complementary line of work evaluates reading comprehension over long or
multi-document inputs. L-Eval \citep{an2024leval} and ZeroSCROLLS
\citep{shaham2023zeroscrolls} require models to process book-length or
multi-document texts, but present these as static reading tasks with no
conversational component. HotpotQA \citep{yang2018hotpotqa} targets multi-hop
reasoning across short Wikipedia passages; it tests inference chains rather than
long-document reading and carries no conversational structure.

\subsection{The Gap}

Table~\ref{tab:benchmarks} summarizes the key dimensions across related
benchmarks. No existing benchmark combines multi-session conversations with long
documents (20K+ tokens) and requires joint reasoning across both.
\mds{} is the first to impose this requirement through the Hybrid source tag.

\begin{table}[t]
  \centering
  \small
  \caption{Comparison of \mds{} with related benchmarks.}
  \label{tab:benchmarks}
  \begin{tabular}{lcccc}
    \toprule
    \textbf{Benchmark} & \textbf{Multi-sess.\ Conv.} & \textbf{Long Docs (20K+)} & \textbf{Joint Reas.} & \textbf{Venue} \\
    \midrule
    LoCoMo \citep{maharana2024evaluating}     & \checkmark & $\times$ & $\times$ & ACL 2024   \\
    LongMemEval \citep{wu2025longmemeval}     & \checkmark & $\times$ & $\times$ & ICLR 2025  \\
    MSC \citep{xu2022beyond}                  & \checkmark & $\times$ & $\times$ & ACL 2022   \\
    MemBench \citep{tan2025membench}          & \checkmark & $\times$ & $\times$ & ACL 2025   \\
    L-Eval \citep{an2024leval}                & $\times$ & \checkmark & $\times$ & ACL 2024   \\
    ZeroSCROLLS \citep{shaham2023zeroscrolls} & $\times$ & \checkmark & $\times$ & EMNLP 2023 \\
    HotpotQA \citep{yang2018hotpotqa}         & $\times$ & $\times$ & $\times$ & EMNLP 2018 \\
    \textbf{Ours (\mds{})}                   & \checkmark & \checkmark & \checkmark & --- \\
    \bottomrule
  \end{tabular}
\end{table}

\subsection{Retrieval-Augmented Generation and Memory Systems}

Retrieval-Augmented Generation (RAG) \citep{lewis2020retrieval} has emerged as a
dominant paradigm for grounding LLM responses in external documents. However,
standard RAG operates over a static document collection indexed at query time,
without modeling the temporal structure of conversations or the navigational
relationship between conversation context and specific documents. Memory systems
such as Mem0 \citep{chhikara2025mem0} and Zep \citep{rasmussen2025zep} augment
LLMs with graph-based or fact-based memory stores that track entities and
relationships across sessions, but are not designed for deep reading within long
documents. Our benchmark exposes both gaps.

% ─────────────────────────────────────────────────────────────────────────────
\section{The \mds{} Benchmark}

\subsection{Task Definition}

We define the task as follows. A system is given a micro-world $M$ consisting of
a set of conversation sessions $C$ and a set of long documents $D$. Given a
natural language question $q$, the system must produce a free-text answer $a$.
Questions may require evidence from $C$ alone, from $D$ alone, or from both---and
the system receives no signal at test time about which source is required.
Evaluation is performed using Exact Match (EM) and token-level F1 against gold
answers, with a separate abstention accuracy metric for adversarial questions.

This formulation is intentionally representation-agnostic: the system may
represent $C$ and $D$ as a flat context, a vector index, a knowledge graph, or
any other structure. The benchmark does not prescribe a retrieval or memory
architecture.

\subsection{Micro-World Structure}

\begin{figure}[t]
  \centering
  \includegraphics[width=0.92\textwidth]{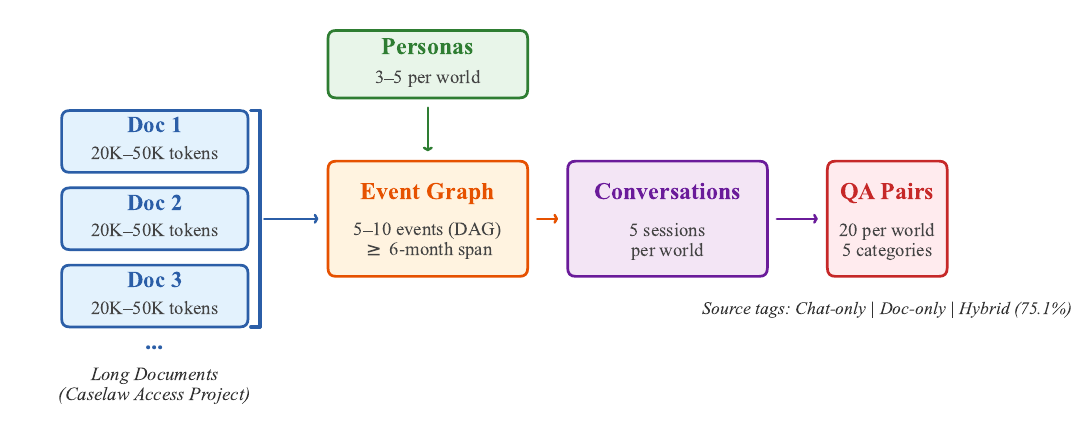}
  \caption{Structure of a \mds{} micro-world. Each micro-world comprises personas,
           a temporal event graph, long documents, multi-session conversations, and
           QA pairs annotated with source tags (Chat-only, Doc-only, Hybrid).}
  \label{fig:microworld}
\end{figure}

The fundamental unit of the dataset is the \textbf{micro-world}---a
self-contained scenario that provides all context needed to answer its associated
questions. Figure~\ref{fig:microworld} illustrates the structure.
Formally, a micro-world is a tuple $M = (P, G, D, C, Q)$.

\paragraph{$P$ (Personas).} $P$ is a set of \textit{personas}, each described
by a name, professional role, domain expertise, communication style, and a set
of directed relationships to other personas (e.g., \textit{reports to},
\textit{collaborates with}). Each micro-world contains 3--5 personas.

\paragraph{$G$ (Event Graph).} $G$ is a \textit{temporal event graph}---a
directed acyclic graph (DAG) over a set of time-stamped events. Each event
$e \in G$ has a timestamp, a natural language description, a subset of personas
involved, and a subset of documents referenced. Events span a minimum of six
months of simulated activity. The DAG structure encodes causal dependencies
between events.

\paragraph{$D$ (Documents).} $D$ is a set of \textit{long documents}, each
containing 20,000--50,000 tokens of real legal text sourced from the Caselaw
Access Project \citep{harvardlaw2018caselaw}. Each micro-world contains 3--5
documents. Documents are shared across personas and events within a micro-world,
grounding the scenario in a consistent body of written evidence.

\paragraph{$C$ (Conversations).} $C$ is a set of \textit{conversation sessions}.
Each session $s \in C$ is anchored to a specific event $e \in G$, involves a
subset of personas from $P$, and consists of a sequence of utterances. Sessions
are ordered chronologically and collectively span the timeline of $G$. At least
40\% of sessions explicitly reference one or more documents in $D$, establishing
the navigational relationship between conversation context and document content
that defines Hybrid questions.

\paragraph{$Q$ (QA Pairs).} $Q$ is a set of \textit{QA pairs}. Each $q \in Q$
consists of a natural language question, a gold free-text answer, a category
label (Section~\ref{sec:categories}), a source tag
(Section~\ref{sec:source-dim}), and one or more evidence references pointing to
the specific utterance or document passage that supports the answer. Each
micro-world contains 20 QA pairs.

\subsection{Question Categories}
\label{sec:categories}

We adopt five question categories following the taxonomy established in LoCoMo
\citep{maharana2024evaluating}, which has become the standard for memory
benchmark evaluation. Table~\ref{tab:categories} defines each category with an
example drawn from the legal domain.

\begin{table}[t]
  \centering
  \small
  \caption{QA categories, definitions, and examples.}
  \label{tab:categories}
  \begin{tabular}{lp{5.2cm}p{6.0cm}}
    \toprule
    \textbf{Category} & \textbf{Definition} & \textbf{Example} \\
    \midrule
    Single-hop      & A single fact retrievable from one source without inference
                    & \textit{``What was the penalty clause amount in the May 15th contract?''} \\[4pt]
    Multi-hop       & Requires chaining two or more facts across sessions or documents
                    & \textit{``Did the delivery deadline in the contract match what was agreed in the March meeting?''} \\[4pt]
    Temporal        & Requires reasoning about the relative or absolute ordering of events
                    & \textit{``Was the settlement offer made before or after the expert witness deposition?''} \\[4pt]
    Knowledge Update & An earlier fact has been superseded by a later one; the correct answer is the most recent
                    & \textit{``What is the current delivery deadline?''} (original: June 1; amended: July 15) \\[4pt]
    Adversarial     & The question contains a false premise or is unanswerable from the provided context
                    & \textit{``What did the NDA signed on May 20th say about non-compete?''} (no NDA signed on May 20th) \\
    \bottomrule
  \end{tabular}
\end{table}

These five categories are not mutually exclusive in general, but in our
annotation each QA pair is assigned exactly one primary category. For Adversarial
questions, we additionally record the adversarial type (\textit{false premise} or
\textit{unanswerable}) and evaluate systems separately on abstention
accuracy---the rate at which a system correctly declines to answer rather than
producing a hallucinated response.

\subsection{The Source Dimension}
\label{sec:source-dim}

The defining contribution of \mds{} is the \textbf{source dimension}: an
orthogonal annotation on every QA pair that identifies which information sources
are required to answer the question. Each QA pair carries exactly one of three
mutually exclusive source tags.

\paragraph{Chat-only.} The gold answer is fully derivable from the conversation
sessions $C$; no document in $D$ needs to be read. This tag tests standard
conversational memory and is comparable to questions in LoCoMo
\citep{maharana2024evaluating} and LongMemEval \citep{wu2025longmemeval}.

\paragraph{Doc-only.} The gold answer requires reading one or more documents in
$D$; the conversation sessions provide no additional signal. This tag tests
long-document comprehension and is comparable to questions in L-Eval
\citep{an2024leval} and ZeroSCROLLS \citep{shaham2023zeroscrolls}.

\paragraph{Hybrid.} Answering requires two steps: first, the system must use
the conversation sessions $C$ to identify \textit{which} document in $D$ is
relevant---typically because a persona references a specific document during a
session; second, it must read that document to extract the answer. Neither step
alone is sufficient. This tag has no equivalent in any prior benchmark.

We enforce that at least 30\% of QA pairs per micro-world carry the Hybrid tag.
Critically, every Hybrid QA pair is verified to satisfy two structural
conditions: (1) at least one session in $C$ contains an explicit reference to the
document required to answer $q$, and (2) the gold answer cannot be derived from
$C$ alone, requiring the system to actually read the referenced document.

\subsection{Dataset Statistics}

\mds{} v1.0 comprises \textbf{50 micro-worlds} in the legal domain (US caselaw
sourced from the Caselaw Access Project), split 70/14/16 into train, validation,
and test sets at the micro-world level (35/7/8 worlds respectively; the slight
asymmetry reflects integer division of 50 whole micro-worlds). Table~\ref{tab:stats} summarizes the
measured structural properties of the released dataset, and
Table~\ref{tab:distribution} shows the QA pair distribution.

\begin{table}[t]
  \centering
  \small
  \caption{\mds{} v1.0 structural properties. Documents are sourced from the
           Caselaw Access Project \citep{harvardlaw2018caselaw}.}
  \label{tab:stats}
  \begin{tabular}{lr}
    \toprule
    \textbf{Property} & \textbf{Value} \\
    \midrule
    Total micro-worlds                  & 50 \\
    Train / Val.\ / Test worlds         & 35 / 7 / 8 \\
    Train / Val.\ / Test QA pairs       & 700 / 140 / 160 \\
    Total QA pairs                      & 1,000 \\
    QA pairs per micro-world            & 20 (uniform) \\
    Total documents (pool)              & 790 \\
    Avg.\ documents per micro-world     & 4.0 (range 3--5) \\
    Document length                     & 20{,}000--50{,}000 tokens \\
    Total conversation sessions         & 250 \\
    Avg.\ sessions per micro-world      & 5.0 \\
    Avg.\ utterances per session        & 10.6 (range 6--16) \\
    Avg.\ personas per micro-world      & 3.9 (range 3--5) \\
    Document source corpus              & Caselaw Access Project \\
    \bottomrule
  \end{tabular}
\end{table}

\begin{table}[t]
  \centering
  \small
  \caption{QA pair distribution by category, source tag, and difficulty.}
  \label{tab:distribution}
  \begin{tabular}{llrr}
    \toprule
    \textbf{Dimension} & \textbf{Label} & \textbf{Count} & \textbf{\%} \\
    \midrule
    \multirow{5}{*}{Category}   & Single-hop     & 200 & 20.0\% \\
                                & Multi-hop      & 200 & 20.0\% \\
                                & Temporal       & 200 & 20.0\% \\
                                & Know.\ Update  & 200 & 20.0\% \\
                                & Adversarial    & 200 & 20.0\% \\
    \midrule
    \multirow{3}{*}{Source tag} & Hybrid         & 751 & 75.1\% \\
                                & Chat-only      & 136 & 13.6\% \\
                                & Doc-only       & 113 & 11.3\% \\
    \midrule
    \multirow{3}{*}{Difficulty} & Medium         & 389 & 38.9\% \\
                                & Hard           & 321 & 32.1\% \\
                                & Easy           & 290 & 29.0\% \\
    \bottomrule
  \end{tabular}
\end{table}

QA pairs are distributed uniformly across the five question categories (20\% each)
by construction. The Hybrid source tag accounts for 75.1\% of all pairs,
substantially exceeding the 30\% design floor, reflecting the fact that most
questions in a scenario grounded on real documents naturally require bridging
conversation and document evidence. The test split (160 pairs, 8 worlds) is the
evaluation set used in all baseline experiments reported in
Section~\ref{sec:experiments}.

\subsection{Quality Analysis}
\label{sec:quality}

We conduct a full-coverage automated quality analysis using an LLM-as-judge
protocol applied to all 50 micro-worlds. Two judge instances are instantiated
from the same model (Claude Sonnet 4.6) with contrasting system prompts: a
\textbf{strict reviewer} instructed to flag any QA pair whose gold answer is not
unambiguously supported by the cited evidence, and a \textbf{lenient reviewer}
instructed to accept answers that are consistent with the evidence under any
reasonable reading. Each judge independently scores every QA pair in a world as
correct or incorrect, producing two binary label vectors per world over which
Cohen's $\kappa$ is computed.

This design measures \textbf{prompt-sensitivity self-consistency}: the degree to
which the pipeline's quality signal is stable across different interpretive
stances toward the same evidence. It is not equivalent to human inter-annotator
agreement---the two judges share a model family with the generation pipeline,
which may introduce leniency bias---but it provides a reproducible, scalable
proxy for labeling consistency.

Across all 50 micro-worlds, the median prompt-sensitivity $\kappa$ is
\textbf{0.634} (mean 0.619, range 0.000--1.000). 22 worlds (44\%) achieve
$\kappa \geq 0.70$ (substantial agreement), 29 worlds (58\%) achieve $\kappa
\geq 0.60$, and 41 worlds (82\%) achieve $\kappa \geq 0.40$. All 50 worlds are
retained in the released dataset; the per-world $\kappa$ values are published
alongside the data to allow downstream users to apply their own quality filters.
Worlds with low $\kappa$ predominantly arise from a systematic base-rate gap
between the two prompts---the strict judge accepts roughly 40\% of QA pairs per
world while the lenient judge accepts roughly 55\%---rather than from disagreement
about specific pairs. This base-rate effect mathematically caps achievable
$\kappa$ for worlds where the gap is large, independent of actual QA correctness.
We discuss this limitation further in Section~\ref{sec:discussion-human}.

% ─────────────────────────────────────────────────────────────────────────────
\section{Collection Pipeline}
\label{sec:pipeline}

\mds{} is generated by an automated seven-stage pipeline that combines real
document sourcing with LLM-driven synthetic structure generation. The pipeline is
fully reproducible, config-driven, and supports checkpoint/resume to handle long
runs.

\subsection{Overview}

\begin{figure}[t]
  \centering
  \includegraphics[width=\textwidth]{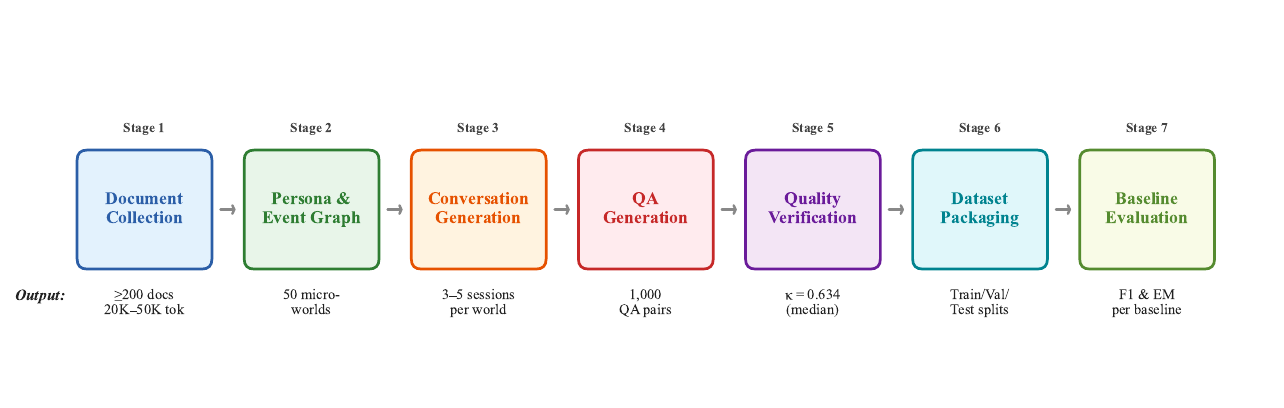}
  \caption{Seven-stage generation and evaluation pipeline. Each stage is
           independently reproducible; the pipeline supports checkpoint/resume
           for long runs.}
  \label{fig:pipeline}
\end{figure}

The seven stages are: (1) Document Collection, (2) Persona \& Event Graph
Generation, (3) Conversation Generation, (4) QA Generation, (5) Quality
Verification, (6) Dataset Packaging, and (7) Baseline Evaluation. All generation
calls in Stages 1--5 use Claude Sonnet 4.6 via the Anthropic API. All LLM calls
are routed through a unified client supporting multiple providers (Groq, Ollama,
Anthropic, OpenAI), making the pipeline provider-agnostic.
Figure~\ref{fig:pipeline} provides an overview of the full pipeline with
per-stage outputs.

\subsection{Stage 1: Document Collection}

Long documents are sourced from the \textbf{Caselaw Access Project}
\citep{harvardlaw2018caselaw}, a Harvard Law School initiative that has digitized
6.9 million US court opinions spanning all jurisdictions from the founding era to
approximately 2020. Individual case opinions typically range from 5,000 to 20,000
characters. We bundle full reporter volumes into single documents, reaching our
20,000--50,000 token target. Documents are validated using the
\texttt{cl100k\_base} tokenizer and must carry a permissive license. The current
corpus covers 10 US jurisdictions with planned expansion to all 50 states.

\subsection{Stage 2: Persona and Event Graph Generation}

For each micro-world, an LLM generates a set of 3--5 personas and a temporal
event graph grounded on the collected documents. The event graph is a DAG of
5--10 time-stamped events, each referencing one or more personas and one or more
documents, spanning a minimum of six months of simulated activity. Event graphs
are validated for structural correctness (no cycles, all referenced entities
exist) before advancing.

\subsection{Stage 3: Conversation Generation}

An LLM generates 5 dialogue sessions for each micro-world, grounded on the event
graph. Each session is anchored to a specific event and involves a subset of the
micro-world's personas. A grounding ratio of at least 40\% of sessions must
reference at least one document, enforcing that conversations are not purely
self-contained.

\subsection{Stage 4: QA Generation}

QA pairs are generated across five categories adapted from the LoCoMo taxonomy
\citep{maharana2024evaluating}: single-hop, multi-hop, temporal, knowledge
update, and adversarial. Each QA pair is additionally annotated with a source
tag; at least 30\% of QA pairs per micro-world must be Hybrid. Gold answers
include evidence references pointing to the specific conversation turn or document
passage that supports the answer.

\subsection{Stage 5: Quality Verification}

All 50 micro-worlds are assessed using the LLM-as-judge protocol described in
Section~\ref{sec:quality}. The per-world $\kappa$ values are recorded and
published with the dataset release. All worlds are retained regardless of $\kappa$
value (\texttt{agreement\_threshold: 0.0}); $\kappa$ is reported as a
prompt-sensitivity quality characterisation rather than a hard filter (see
Section~\ref{sec:quality} for rationale).

\subsection{Stage 6: Dataset Packaging}

Verified micro-worlds are split 70/14/16 into train, validation, and test sets at
the micro-world level. Splitting is performed with document isolation: no document
appears in more than one split, preventing leakage. The dataset is serialized to
JSON with accompanying statistics and a Datasheets for Datasets
\citep{gebru2021datasheets} documentation file.

% ─────────────────────────────────────────────────────────────────────────────
\section{Baseline Experiments}
\label{sec:experiments}

We evaluate six baseline configurations that span the design space of current
approaches to conversational memory and long-document retrieval. These baselines
are intended to establish lower and upper bounds on the benchmark and to
characterize where each class of approach succeeds and fails---not to achieve
state-of-the-art performance.

\subsection{Experimental Setup}

All baselines use \textbf{Claude Sonnet 4.5} for answer generation. The dataset
itself was generated with Claude Sonnet 4.6, keeping generation and evaluation
models distinct to avoid self-evaluation bias. Documents are chunked into
512-token windows with 64-token overlap. Chunk embeddings are computed with the
\texttt{all-MiniLM-L6-v2} sentence encoder \citep{reimers2019sentence} and
indexed in a persistent ChromaDB \citep{chroma2022chroma} vector store. Retrieval uses cosine similarity with $k{=}10$ chunks; this value was not tuned
on the validation split and hyperparameter sensitivity is left to future work.
All experiments are conducted on the test split only (160 QA pairs, 8
micro-worlds); the train and validation splits are reserved for system
development.

\subsection{Baselines}

We define six baseline configurations, summarized in Table~\ref{tab:baselines}.

\begin{table}[t]
  \centering
  \small
  \caption{Baseline configurations.}
  \label{tab:baselines}
  \begin{tabular}{llll}
    \toprule
    \textbf{Baseline} & \textbf{Context Source} & \textbf{Context Size} & \textbf{What It Tests} \\
    \midrule
    Base LLM         & Conv.\ + documents (truncated) & 4,096 tokens    & Lower bound: parametric knowledge + short context \\
    Long-Context LLM & Conv.\ + documents (full)      & 60,000 tokens   & Long-context capacity with no retrieval \\
    RAG-Conv         & Top-$k$ conversation chunks    & $k{=}10$ chunks & Retrieval over conversation history only \\
    RAG-Doc          & Top-$k$ document chunks        & $k{=}10$ chunks & Retrieval over long documents only \\
    RAG-Both         & Top-$k$ chunks from both       & $k{=}10$ chunks & Joint retrieval over both modalities \\
    Memory System    & Mem0-style extracted facts     & Top-$k$ facts   & Graph-based memory augmentation \\
    \bottomrule
  \end{tabular}
\end{table}

\paragraph{Base LLM} concatenates all conversation sessions and document text for
the relevant micro-world and truncates to a 4,096-token context window. This
serves as a lower bound, since it can only use whatever fits in the window.

\paragraph{Long-Context LLM} provides the full concatenation of conversation
sessions and document text up to a 60,000-token context limit. It tests whether a
long-context model can solve the benchmark purely through extended attention.

\paragraph{RAG-Conv} indexes only conversation utterances. At inference time, the
top-$k$ utterance chunks most similar to the question are retrieved. This
configuration is directly comparable to retrieval-augmented approaches used in
existing memory benchmarks \citep{maharana2024evaluating,wu2025longmemeval}, and
has no access to document content.

\paragraph{RAG-Doc} indexes only document chunks. This is the natural RAG
approach for long-document QA benchmarks \citep{an2024leval,shaham2023zeroscrolls}
and serves as a strong Doc-only baseline. It has no access to conversation
history.

\paragraph{RAG-Both} indexes both conversation utterances and document chunks into
a unified vector store and retrieves the top-$k$ chunks across both modalities,
ranked jointly by similarity score. This is the strongest retrieval baseline.

\paragraph{Memory System} implements a fact-extraction memory layer on top of
ChromaDB \citep{chroma2022chroma}. For each conversation session, an LLM prompt
extracts up to 10 key facts as a structured list; these facts are embedded and
stored in a dedicated vector collection. At query time, the top-$k$ most similar
facts are retrieved as context. Unlike the RAG baselines, which operate on raw
text chunks, the Memory System stores semantic abstractions of conversations.
Importantly, conversation sessions frequently reference and paraphrase document
content, so the extracted facts can capture document-derived information
indirectly---explaining the system's competitive Doc-only score despite having no
direct access to document chunks.

\subsection{Evaluation Metrics}

\paragraph{Exact Match (EM)} A prediction is correct if, after normalization
(lowercasing, removing articles and punctuation), it exactly matches the gold
answer string.

\paragraph{Token-level F1} The harmonic mean of precision and recall computed
over the bag of tokens in the predicted and gold answers, after the same
normalization.

\paragraph{Abstention Accuracy} For Adversarial questions only, we evaluate
whether the system correctly abstains rather than producing a hallucinated answer.
A system is instructed to respond with the special token \texttt{ABSTAIN} when it
cannot answer from the provided context. This metric is reported separately and
not included in overall EM or F1.

Token-level F1 and Abstention Accuracy are reported overall and broken down
along two dimensions: (1) question category and (2) source tag. Exact Match
is omitted from the main tables as free-text answers rarely produce exact
string matches; F1 is the primary metric throughout. The per-source-tag F1
breakdown is the primary diagnostic for measuring the joint reasoning gap.

\subsection{Results}

Tables~\ref{tab:results-source} and \ref{tab:results-category} report F1 results
on the test split.

\begin{table}[t]
  \centering
  \small
  \caption{Token-level F1 per baseline by source tag (test split, $n{=}160$).
           Abstention Accuracy (AbstAcc) is reported separately for Adversarial
           questions ($n{=}32$). Bold indicates the highest value in each column.
           Test-split counts for Doc-only ($n{=}23$) and Chat-only ($n{=}16$) are
           small; per-tag differences should be interpreted with appropriate
           caution regarding statistical power.}
  \label{tab:results-source}
  \begin{tabular}{lccccc}
    \toprule
    \textbf{Baseline} & \textbf{Hybrid} ($n{=}121$) & \textbf{Doc-Only} ($n{=}23$) & \textbf{Chat-Only} ($n{=}16$) & \textbf{Overall} ($n{=}160$) & \textbf{AbstAcc} \\
    \midrule
    Base LLM         & 0.317 & 0.339          & 0.285          & 0.317          & 0.938          \\
    Long-Context LLM & 0.330 & 0.318          & 0.283          & 0.323          & 0.844          \\
    RAG-Conv         & \textbf{0.348} & 0.351          & 0.275          & 0.341          & 0.938          \\
    RAG-Doc          & 0.267          & 0.453          & 0.292          & 0.296          & \textbf{1.000} \\
    RAG-Both         & 0.342          & \textbf{0.475} & \textbf{0.315} & \textbf{0.358} & 0.969          \\
    Memory System    & 0.302 & 0.459          & 0.309          & 0.325          & 0.969          \\
    \bottomrule
  \end{tabular}
\end{table}

\begin{table}[t]
  \centering
  \small
  \caption{Token-level F1 per baseline by question category (test split,
           $n{=}160$, 32 per category).}
  \label{tab:results-category}
  \begin{tabular}{lccccc}
    \toprule
    \textbf{Baseline} & \textbf{Single-Hop} & \textbf{Multi-Hop} & \textbf{Temporal} & \textbf{Know.\ Update} & \textbf{Adversarial} \\
    \midrule
    Base LLM         & 0.331          & 0.318          & 0.319          & 0.314          & 0.304          \\
    Long-Context LLM & 0.310          & 0.310          & 0.340          & 0.329          & 0.327          \\
    RAG-Conv         & 0.347          & \textbf{0.324} & 0.334          & \textbf{0.372} & 0.329          \\
    RAG-Doc          & 0.408          & 0.277          & 0.220          & 0.233          & 0.342          \\
    RAG-Both         & \textbf{0.431} & 0.298          & \textbf{0.345} & 0.367          & \textbf{0.350} \\
    Memory System    & 0.406          & 0.269          & 0.305          & 0.321          & 0.327          \\
    \bottomrule
  \end{tabular}
\end{table}

The most striking pattern in Table~\ref{tab:results-source} is the
\textbf{asymmetry of specialization}. RAG-Doc achieves the second-highest F1 on
Doc-only questions (0.453) but the \textit{lowest} overall score (0.296) due to
near-collapse on Hybrid questions (0.267). This directly demonstrates that
document retrieval without conversational navigation is insufficient for joint
reasoning. RAG-Both recovers this gap, improving Hybrid F1 from 0.267 to 0.342
by incorporating conversation retrieval alongside document retrieval.

Long-Context LLM ranks fourth overall (0.323) despite receiving the full
60,000-token context window. Its comparatively low abstention accuracy (0.844
vs.\ 0.938--1.000 for retrieval baselines) indicates that attending over very
long contexts increases the tendency to fabricate answers for unanswerable
questions. The Memory System (0.325 overall) is competitive with the long-context
approach, particularly on Doc-only questions (0.459), where its fact-extraction
layer distils document-referenced facts from conversations into easily retrievable
form.

In Table~\ref{tab:results-category}, all baselines show relatively uniform F1
across categories (range 0.267--0.431), with the exception of RAG-Doc which
degrades sharply on Temporal (0.220) and Knowledge Update (0.233)
questions---categories requiring tracking evolving state across sessions,
information unavailable to document-only retrieval. Multi-hop questions are
consistently the weakest category for retrieval baselines (RAG-Both: 0.298),
suggesting that chaining evidence across two or more retrieval steps remains an
open challenge.

% ─────────────────────────────────────────────────────────────────────────────
\section{Discussion}

\subsection{Do Hybrid Questions Expose a Genuine Joint-Retrieval Gap?}

The answer is partially yes, but with important nuance. For baselines that
specialize on one modality, Hybrid is clearly the hardest source tag: RAG-Doc
scores 0.267 on Hybrid versus 0.453 on Doc-only, and Memory System scores 0.302
on Hybrid versus 0.459 on Doc-only. The pattern is consistent---a system that
lacks access to one modality is penalized precisely on the questions that require
it.

However, the gap is less pronounced for baselines that have access to all
context. Long-Context LLM actually scores \textit{highest} on Hybrid (0.330)
among the three source tags, outperforming its own Doc-only score (0.318), likely
because Hybrid questions tend to have richer evidence in the conversation turns
and because the question text echoes conversational phrasing. The specialization
asymmetry---strong on one source tag, weak on the complementary tag---is the
clearest signal that the benchmark is measuring something real.

\subsection{Does RAG-Both Improve Over Specialised Baselines on Hybrid?}

RAG-Both (Hybrid F1=0.342) substantially outperforms RAG-Doc (0.267) on Hybrid
questions, confirming that incorporating conversation retrieval recovers most of
the gap caused by document-only indexing. However, RAG-Both does not
substantially outperform RAG-Conv (0.348) on Hybrid questions. This is a striking
finding: on Hybrid questions, conversation-side retrieval appears to carry most of
the weight. The likely explanation is that Hybrid question text tends to echo
conversation language (persona names, event references) more closely than it
echoes specific document terminology, making conversation chunks more similar to
queries in embedding space.

This result suggests that the joint retrieval challenge in Hybrid questions is not
symmetric: the hard part is identifying \textit{which document} a conversation
references (a navigation problem), not retrieving passages from that document once
it is identified. No baseline we evaluate implements this two-stage strategy.

\subsection{Does Long-Context LLM Outperform Base LLM?}

Despite a 15$\times$ context size increase (4,096 $\to$ 60,000 tokens),
Long-Context LLM achieves only a marginal overall improvement over Base LLM
(0.323 vs.\ 0.317 F1). On Doc-only questions, Long-Context LLM is actually
\textit{worse} than Base LLM (0.318 vs.\ 0.339), and its abstention accuracy is
the lowest of all baselines (0.844 vs.\ 0.938 for Base LLM). This pattern is
consistent with the ``lost in the middle'' failure mode documented for long-context
transformers \citep{liu2024lost}, in which relevant content embedded deep within a
long context is systematically underweighted by attention. The practical
implication is that expanding the context window is not a substitute for
structured retrieval. Systems with access to the full context but no retrieval
mechanism do not substantially outperform systems that see only 4,096 tokens of
it---while simultaneously becoming less reliable on adversarial questions. Taken
together, the results across all six baselines suggest that neither context length
nor fact-based memory substitutes for structured joint retrieval---the core design
challenge this benchmark is intended to expose.

\subsection{Human-System Gap}
\label{sec:discussion-human}

A formal human performance upper bound has not been measured in this work and
remains an important direction for future evaluation. Recruiting
domain-knowledgeable annotators for caselaw material, ensuring sufficient
annotation time for long-document comprehension, and controlling for individual
variance all require infrastructure beyond the current scope.

We anticipate the human-system gap to be substantial, particularly on Hybrid
questions. As a partial quality signal, our LLM-as-judge verification
(Section~\ref{sec:quality}) found that the lenient judge accepted 67\% of QA
pairs overall, with higher acceptance on Doc-only (96\%) and Chat-only (61\%)
questions than on Hybrid (44\%). The lower Hybrid acceptance rate is consistent
with the difficulty observed in baseline F1 scores. Future work reporting human F1
on a sample of $\geq$50 test questions would anchor the benchmark's difficulty and
provide a concrete target for system improvement.

\subsection{Implications for System Design}

The results collectively point to a specific architectural gap. No existing
approach---long-context attention, retrieval over one modality, retrieval over
both, or fact-based memory---implements the two-step process that Hybrid questions
require: (1) use conversation context to identify \textit{which} document is
relevant, and (2) retrieve specifically from that document. RAG-Both comes closest
but does so with a flat similarity ranking that cannot enforce the navigational
dependency between step 1 and step 2.

A system that could close the Hybrid gap would need at minimum: (a) a
representation of which conversation sessions reference which documents (a
citation graph), and (b) a retrieval strategy that conditions document-chunk
retrieval on the output of conversation-level retrieval. Memory systems designed
around entity and relationship graphs (e.g., Mem0 \citep{chhikara2025mem0}, Zep
\citep{rasmussen2025zep}) could in principle encode document references as graph
edges, but neither system in its current form is designed for this use case.
\mds{} makes the gap measurable and provides the scaffolding for evaluating future
approaches that attempt to close it.

% ─────────────────────────────────────────────────────────────────────────────
\section*{Limitations}

\textbf{Domain scope.} The current document corpus is drawn entirely from US
court opinions sourced from the Caselaw Access Project
\citep{harvardlaw2018caselaw}. While legal text is a natural fit for the
benchmark, it introduces domain bias. We plan to expand to additional document
types and domains in future releases.

\textbf{LLM-generated conversation quality.} Conversation sessions are generated
by a language model conditioned on the event graph and document set. LLM-generated
dialogues may not capture the full range of natural language phenomena found in
real human conversations---including implicit references, pragmatic inferences,
code-switching, and repair sequences.

\textbf{Automated QA generation.} QA pairs and gold answers are produced by an
LLM and validated through structural checks. Automated generation cannot guarantee
that every gold answer is uniquely correct, that distractors in Adversarial
questions are sufficiently realistic, or that Knowledge Update questions capture
all intermediate states.

\textbf{LLM-as-judge verification limitations.} Quality verification is performed
by two LLM judge instances from the same model family as the generation pipeline,
introducing potential self-leniency bias. Additionally, the prompt-sensitivity
$\kappa$ we measure reflects base-rate differences between strict and lenient
prompts as much as genuine label disagreement.

\textbf{English only.} The pipeline and document corpus are currently
English-only.

\textbf{Synthetic scenario realism.} Micro-world personas, event graphs, and
conversation sessions are synthetically generated. The temporal event graphs
enforce a clean DAG structure, which may be more regular than the overlapping,
ambiguous timelines found in real cases.

\textbf{Fixed Hybrid threshold.} The 30\% Hybrid question floor is a dataset
design parameter chosen to ensure reliable per-source-tag metrics. It is not
empirically derived from a study of real-world task distributions.

% ─────────────────────────────────────────────────────────────────────────────
\section{Conclusion}

We introduced \mds{} to close a gap that existing benchmarks leave open: no prior
work evaluates AI systems on the joint task of navigating multi-session
conversation history \textit{and} performing deep reading within long documents
simultaneously. The Hybrid source tag formalizes this requirement---questions that
are unanswerable without first using conversation context to identify the relevant
document, then reading that document to extract the answer.

Our baseline experiments reveal a clear \textbf{specialization asymmetry} with
direct implications for system design. RAG-Doc, the natural approach for
long-document QA, collapses to 0.267 F1 on Hybrid despite achieving 0.453 on
Doc-only---a 41\% relative drop caused by the structural inability of
document-only retrieval to resolve which document a conversation is referring to.
RAG-Both closes much of this gap (Hybrid F1=0.342) but the remaining shortfall
points to the core open problem: no existing baseline implements the two-step
navigational strategy that Hybrid questions require. A system that first uses
conversation retrieval to identify the relevant document, then applies targeted
retrieval within that specific document, would be architecturally suited to this
task---and no current approach does so.

The practical implication is that the next generation of AI assistants will need
to couple conversational memory with long-document navigation at the architectural
level, not just at the context-window level. \mds{} makes this gap measurable,
reproducible, and quantifiable for the first time. We release the dataset,
generation pipeline, and all baseline implementations to support this line of
research.

% ─────────────────────────────────────────────────────────────────────────────
% Manual bibliography: each entry shows authors, title (clickable -> hidden DOI),
% venue/journal, volume/issue/pages, year. Order = citation order ([1]--[15]).

% ─────────────────────────────────────────────────────────────────────────────
\appendix

\section*{Appendix}

\section{Pipeline Stage Details}
\label{app:pipeline}

\paragraph{Stage 1 --- Document Collection.} Raw opinions are downloaded from the
Caselaw Access Project bulk export and processed into long document objects. Each
document is tokenized with \texttt{cl100k\_base} and filtered to 20,000--50,000
tokens. Documents are deduped by case ID and split into per-state shards to
support reproducible batched runs.

\paragraph{Stage 2 --- Persona \& Event Graph Generation.} For each micro-world
the pipeline selects 3--5 documents at random (without replacement), infers a
domain label from their content, and calls the persona and event-graph generators
in sequence. Persona count is sampled from \texttt{personas\_per\_world} (default
3--5). Event count is sampled from \texttt{events\_per\_graph} (default 5--10).
The event graph is validated as a DAG before acceptance; malformed graphs are
retried up to three times.

\paragraph{Stage 3 --- Conversation Generation.} Sessions are generated one per
event in topological order of the event graph, so that each session can reference
prior-session context. A grounding flag controls whether a session must include at
least one document citation (applied to \texttt{grounding\_ratio\_threshold}
$\times$ sessions). Each session is generated independently, making the stage
trivially parallelisable.

\paragraph{Stage 4 --- QA Generation.} Five category-specific prompts are called
per micro-world. A post-generation filter enforces at least
\texttt{min\_qa\_per\_category} pairs per category and at least
\texttt{doc\_navigation\_ratio} pairs with \texttt{requires\_doc\_navigation=true}.
Pairs failing structural validation are discarded; the stage retries if minimum
counts are not met.

\paragraph{Stage 5 --- Quality Verification.} Two LLM judge instances (strict and
lenient system prompts, same base model) evaluate every micro-world in parallel.
Each judge returns per-QA correctness labels (\texttt{qa\_answer\_correctness})
and per-world Likert quality scores (persona consistency, temporal coherence,
document grounding accuracy, 1--5). Cohen's $\kappa$ is computed over the
\texttt{qa\_answer\_correctness} labels. All micro-worlds are retained regardless
of $\kappa$ value (\texttt{agreement\_threshold: 0.0}); $\kappa$ is reported as a
prompt-sensitivity quality characterisation rather than a hard filter (see
Section~\ref{sec:quality} for rationale).

\paragraph{Stage 6 --- Dataset Packaging.} Verified micro-worlds are shuffled
(seed fixed to \texttt{random\_seed}) and split 70/14/16 into train, validation,
and test sets at the micro-world level. No QA pair from one split shares a
micro-world with another split. Document-level leakage is also prevented: a
document assigned to a micro-world in the test split does not appear in any train
or validation micro-world.

\paragraph{Stage 7 --- Baseline Evaluation.} Each baseline system is given the
question text and the access mode appropriate to its design. Predictions are
scored against gold answers using token-level F1 and exact match after
lowercasing, article removal, and punctuation stripping. Abstention accuracy is
computed separately for Adversarial questions using a reserved \texttt{ABSTAIN}
token.

% ─────────────────────────────────────────────────────────────────────────────
\section{Prompt Templates}
\label{app:prompts}

All LLM calls use a two-part message: a system prompt setting the generator's
role and a user prompt containing the structured input and output schema.
Templates are reproduced verbatim below; slot names in \texttt{\{braces\}} are
filled at runtime.

\subsection{Persona Generation}

\begin{tcolorbox}[promptstyle, title=System prompt]
You are a creative writer specializing in generating realistic professional
personas for synthetic dataset generation. You produce well-structured JSON
output that strictly follows the requested schema.
\end{tcolorbox}

\begin{tcolorbox}[promptstyle, title=User prompt]
Generate \texttt{\{count\}} unique professional personas for a micro-world in
the ``\texttt{\{domain\}}'' domain. Each persona must have: a unique full name;
a professional role relevant to the domain; an expertise domain; a communication
style; relationships to other personas. Return a JSON array where each element
has: \texttt{name}, \texttt{role}, \texttt{expertise\_domain},
\texttt{communication\_style}, \texttt{relationships}. Document summaries for
context: \texttt{\{document\_summaries\}}. Return ONLY valid JSON, no markdown
fences or extra text.
\end{tcolorbox}

\subsection{Event Graph Generation}

\begin{tcolorbox}[promptstyle, title=System prompt]
You are an expert at creating realistic temporal event sequences for professional
scenarios. You produce well-structured JSON output that strictly follows the
requested schema. Events must form a valid directed acyclic graph (DAG) with no
cycles.
\end{tcolorbox}

\begin{tcolorbox}[promptstyle, title=User prompt]
Generate a temporal event graph with \texttt{\{event\_count\}} events for a
``\texttt{\{domain\}}'' micro-world. Events must span at least
\texttt{\{time\_span\_months\}} months, form a valid DAG, reference at least one
persona and one document each, and ensure every document is referenced by at
least one event. Return a JSON object with: \texttt{time\_range\_start},
\texttt{time\_range\_end}, and an \texttt{events} array. Each event has:
\texttt{event\_id}, \texttt{timestamp} (ISO 8601), \texttt{description},
\texttt{persona\_ids}, \texttt{document\_ids}, \texttt{depends\_on}. Return
ONLY valid JSON, no markdown fences or extra text.
\end{tcolorbox}

\subsection{Conversation Session Generation}

\begin{tcolorbox}[promptstyle, title=System prompt]
You are an expert dialogue writer who creates realistic multi-party professional
conversations. You produce well-structured JSON output that strictly follows the
requested schema. Conversations must be grounded in the provided event context
and reference documents when appropriate.
\end{tcolorbox}

\begin{tcolorbox}[promptstyle, title=User prompt]
Generate a realistic conversation session for the following event in a
``\texttt{\{domain\}}'' scenario. Event: \texttt{\{event\_description\}} at
\texttt{\{event\_timestamp\}}. Participating personas:
\texttt{\{personas\_json\}}. Documents available for reference:
\texttt{\{documents\_json\}}. Prior conversation context:
\texttt{\{prior\_context\}}. Instructions: generate natural dialogue; respect
each persona's communication style; include temporal markers;
\texttt{\{grounding\_instruction\}}; aim for \texttt{\{target\_utterances\}}
utterances. Return a JSON object with an \texttt{utterances} array and
\texttt{referenced\_document\_ids}. Return ONLY valid JSON.
\end{tcolorbox}

\subsection{QA Generation}

All five category prompts share the same system prompt:

\begin{tcolorbox}[promptstyle, title=System prompt (shared across all five categories)]
You are an expert benchmark dataset creator who generates high-quality
question-answer pairs for evaluating AI systems on joint conversational memory
and long document reasoning. You produce well-structured JSON output that
strictly follows the requested schema.
\end{tcolorbox}

Category-specific core instructions are summarized in Table~\ref{tab:qa-prompts}.
All category prompts request the same output schema per QA pair:
\texttt{question}, \texttt{gold\_answer}, \texttt{evidence\_references} (with
\texttt{source\_type}, \texttt{source\_id}, \texttt{passage\_span}),
\texttt{difficulty}, \texttt{requires\_doc\_navigation}.

\begin{table}[H]
  \centering
  \small
  \caption{QA generation category-specific instructions.}
  \label{tab:qa-prompts}
  \begin{tabular}{lp{5cm}}
    \toprule
    \textbf{Category} & \textbf{Core instruction} \\
    \midrule
    Single-hop    & Each question answerable from a single source; some require navigating from conversation to document. \\
    Multi-hop     & Each question requires facts from $\geq$2 distinct sources; chain reasoning across sessions or across conversation and document. \\
    Temporal      & Each question requires reasoning about chronological ordering; reference $\geq$2 distinct events/timestamps. \\
    Know.\ Update & Each question tests use of the most current information; at least one superseded fact must exist. \\
    Adversarial   & Mix of \texttt{false\_premise}, \texttt{unanswerable}, and \texttt{abstain} types. \\
    \bottomrule
  \end{tabular}
\end{table}

\subsection{LLM-as-Judge (Quality Verification)}

Two instances of the same model are run with contrasting system prompts and an
identical user prompt body.

\begin{tcolorbox}[promptstyle, title=Strict system prompt]
You are a strict quality reviewer for a benchmark dataset of QA pairs grounded
in conversations and long documents. Apply skeptical scrutiny: if a QA pair's
gold answer is not unambiguously supported by the cited evidence, mark it
incorrect. If a question's category label is debatable, mark it inaccurate.
Err on the side of flagging. Output a single JSON object matching the requested
schema. Do not add commentary outside the JSON.
\end{tcolorbox}

\begin{tcolorbox}[promptstyle, title=Lenient system prompt]
You are a charitable quality reviewer for a benchmark dataset of QA pairs
grounded in conversations and long documents. Apply reasonable interpretation:
if the gold answer is consistent with the cited evidence under any reasonable
reading, mark it correct. Only flag clear errors or unsupported claims. Output
a single JSON object matching the requested schema. Do not add commentary
outside the JSON.
\end{tcolorbox}

\paragraph{User prompt (shared).} A compact context is rendered per micro-world
containing the list of personas, the full conversation sessions, and each QA
pair with its cited evidence excerpts inlined. The judge returns a JSON object
with \texttt{persona\_consistency} (1--5), \texttt{temporal\_coherence} (1--5),
\texttt{document\_grounding\_accuracy} (1--5), \texttt{qa\_answer\_correctness}
(map of QA ID $\to$ bool), \texttt{qa\_category\_accuracy} (map of QA ID
$\to$ bool), \texttt{flagged\_qa\_ids}, and \texttt{notes}.

% ─────────────────────────────────────────────────────────────────────────────
\section{Dataset Example}
\label{app:example}

Below is an illustrative excerpt from one micro-world in the caselaw domain (all
names and case references are from authentic Caselaw Access Project documents).

\begin{tcolorbox}[promptstyle, title={Personas (excerpt)}]
\begin{itemize}[leftmargin=*]
  \item \textit{Margaret Chen} --- Senior Partner; contract law; ``formal and precise''
  \item \textit{David Rodriguez} --- Associate Attorney; litigation; ``analytical and thorough''
  \item \textit{Sarah Kim} --- Legal Researcher; case law research; ``methodical and detail-oriented''
\end{itemize}
\end{tcolorbox}

\begin{tcolorbox}[promptstyle, title={Event graph (excerpt, 3 of 7 events)}]
\begin{itemize}[leftmargin=*]
  \item \texttt{evt-1} (2024-01-10): Initial client consultation on breach of contract; involves Chen, Rodriguez; references Case A.
  \item \texttt{evt-3} (2024-02-14): Discovery of precedent contradicting client position; involves Kim; references Case B.
  \item \texttt{evt-5} (2024-03-22): Strategy revision meeting; involves all three personas; references Cases A and B.
\end{itemize}
\end{tcolorbox}

\begin{tcolorbox}[promptstyle, title={QA example --- Single-hop}]
\textbf{Q:} Which attorney conducted the initial client consultation on the breach of contract claim?\\
\textbf{A:} Margaret Chen and David Rodriguez.\\
\textbf{Source:} \texttt{evt-1} conversation session.
\end{tcolorbox}

\begin{tcolorbox}[promptstyle, title={QA example --- Hybrid (doc navigation required)}]
\textbf{Q:} What specific contractual clause did the precedent case identified by Kim in February establish as unenforceable?\\
\textbf{A:} [answer drawn from Case B document after identifying it via \texttt{evt-3} conversation]\\
\textbf{Source:} \texttt{evt-3} conversation session $\to$ Case B document.
\end{tcolorbox}

\begin{tcolorbox}[promptstyle, title={QA example --- Knowledge Update}]
\textbf{Q:} What was the legal team's recommended strategy for the breach of contract case as of March 2024?\\
\textbf{A:} [revised strategy from \texttt{evt-5}, superseding the initial strategy from \texttt{evt-1}]\\
\textbf{Source:} \texttt{evt-5} conversation session; \texttt{evt-1} (superseded).
\end{tcolorbox}

\end{document}